\documentclass{article}
\usepackage{spconf,amsmath,graphicx,hyperref}
\usepackage{cite}
\usepackage{multirow}
\usepackage{amsmath,amssymb,amsfonts}
\usepackage{algorithmic}
\usepackage{graphicx}
\usepackage{subcaption}
\usepackage{caption}
\usepackage{textcomp}
\usepackage{booktabs} 
\usepackage{graphicx}        
\usepackage{booktabs}        
\usepackage{multirow}        
\usepackage[dvipsnames,table]{xcolor}

\usepackage{xcolor}
\usepackage{hyperref} 
\usepackage{tikz}
\usetikzlibrary{arrows.meta,positioning,fit,calc,shadows.blur}

\usepackage{enumitem}

\usepackage{url}

\usepackage{hyperref}

\title{Scale-Aware Self-Supervised Learning for Segmentation of Small and Sparse Structures}
\name{Jorge Quesada and Ghassan AlRegib\thanks{This work is supported by the ML4Seismic Industry Partners at Georgia Tech}}
\address{OLIVES at the Georgia Institute of Technology}

\begin{document}


\onecolumn 

\begin{description}[labelindent=0cm,leftmargin=3cm,style=multiline]

\item[\textbf{Citation}]{J. Quesada, and G. AlRegib, "Scale-Aware Self-Supervised Learning for Segmentation of Small and Sparse Structures", accepted at  2026 IEEE International Conference on Acoustics, Speech, and Signal Processing (ICASSP)}

\item[\textbf{Review}]{Date of acceptance: 17 January 2026}

\item[\textbf{Code \& Data}]{\url{https://github.com/olivesgatech/SASSL}}



\item[\textbf{Bib}]{@ARTICLE\{Quesada2026Scale,\\ 
author=\{J. Quesada and G. AlRegib\},\\ 
journal=\{IEEE International Conference on Acoustics, Speech, and Signal Processing\},\\ 
title=\{Scale-Aware Self-Supervised Learning for Segmentation of Small and Sparse Structures\}, \\ 
year=\{2026\},\}\\ 
}


\item[\textbf{Copyright}]{\textcopyright Creative Commons Attribution CCBY 4.0 }

\item[\textbf{Contact}]{\{jpacora3, alregib\}@gatech.edu  \\ \url{https://alregib.ece.gatech.edu/} }

\item[\textbf{Corresponding author}]{alregib@gatech.edu }

\end{description}

\thispagestyle{empty}
\newpage
\clearpage
\setcounter{page}{1}

\twocolumn


\ninept 
\maketitle

\begin{abstract}
Self-supervised learning (SSL) has emerged as a powerful strategy for representation learning under limited annotation regimes, yet its effectiveness remains highly sensitive to many factors, especially the nature of the target task. In segmentation, existing  pipelines are typically tuned to large, homogeneous regions, but their performance drops when objects are small, sparse, or locally irregular. In this work, we propose a scale-aware SSL adaptation  that integrates small-window cropping into the augmentation pipeline, \texttt{zooming in} on fine-scale structures during pretraining. We evaluate this approach across two domains with markedly different data modalities: seismic imaging, where the goal is to segment sparse faults, and neuroimaging, where the task is to delineate small cellular structures. In both settings, our method yields consistent improvements over standard and state-of-the-art baselines under label constraints, improving accuracy by up to 13\% for fault segmentation and 5\% for cell delineation. In contrast, large-scale features such as seismic facies or tissue regions see little benefit, underscoring that the value of SSL depends critically on the scale of the target objects. Our findings highlight the need to align SSL design with object size and sparsity, offering a general principle for buil ding more effective representation learning pipelines across scientific imaging domains.
\end{abstract}


\section{Introduction}
\label{sec:introduction}

Self-supervised learning (SSL) has become a cornerstone of modern representation learning, enabling models to leverage large unlabeled datasets through pretext tasks rather than costly manual annotations \cite{chen2020simple, he2020momentum, yeh2022decoupled}. Its effectiveness has not only driven advances across vision \cite{chen2020simple, chen2021exploring}, medical imaging \cite{quesada2022mtneuro, li2024self}, and remote sensing \cite{wang2022self, muhtar2023cmid}, but has also laid the foundation for the rise of transformer architectures \cite{kirillov2023segment, Quesada_2024_CVPR, quesada2024benchmarking}, whose pretraining paradigms are inherently self-supervised \cite{devlin2019bert, caron2021emerging}. Despite these successes, most existing SSL pipelines implicitly assume that learned features should capture broad, homogeneous patterns. This design choice aligns well with tasks such as classification or object recognition, but its suitability for tasks dominated by small or sparse structures remains unclear.

\begin{figure}[t]
    \begin{subfigure}{\linewidth}
    \centering
        \includegraphics[width=\textwidth]{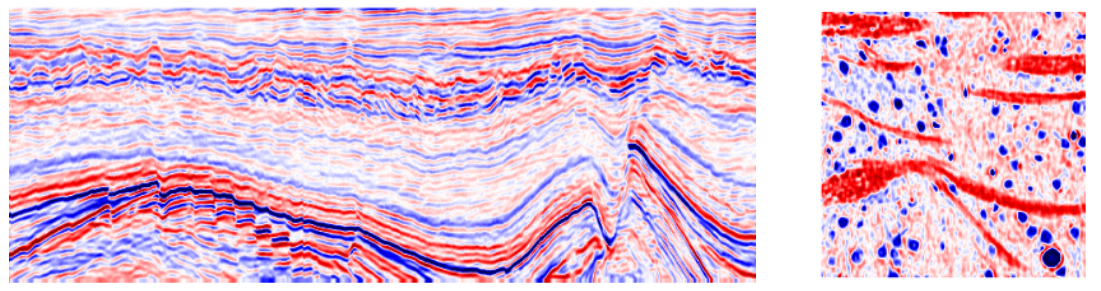} 
        \caption{Seismic and brain image slices}
        \label{fig:mot_raw}
    \end{subfigure}
    \begin{subfigure}{\linewidth}
    \centering
        \includegraphics[width=\textwidth]{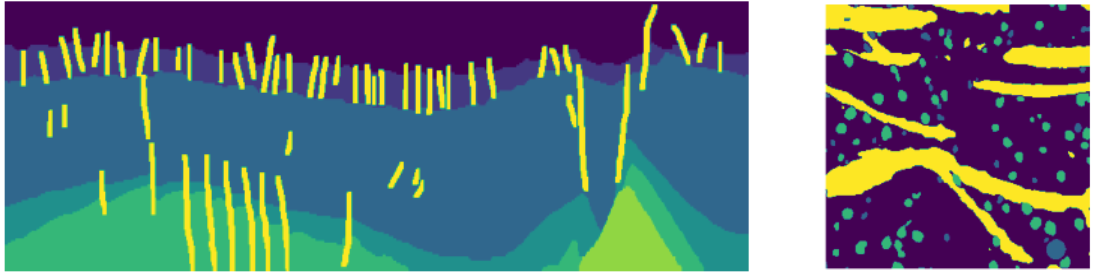} 
        \caption{Ground truth}
        \label{fig:mot_gt}
    \end{subfigure}
    \begin{subfigure}{\linewidth}
        \includegraphics[width=\textwidth]{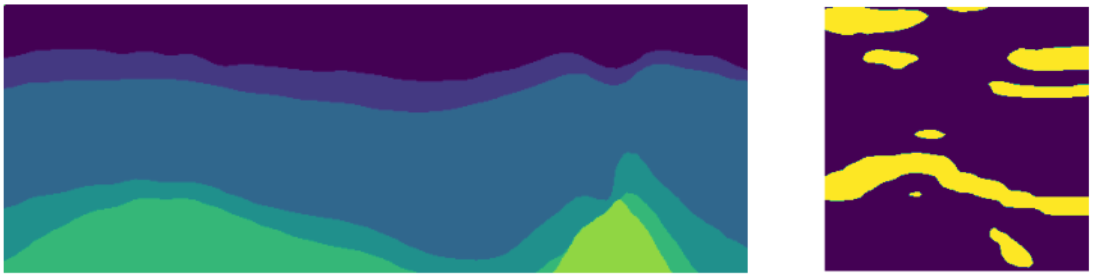}  
        \caption{SSL segmentation of large structures}
        \label{fig:mot_fac}
    \end{subfigure}
    \begin{subfigure}{\linewidth}
        \includegraphics[width=\textwidth]{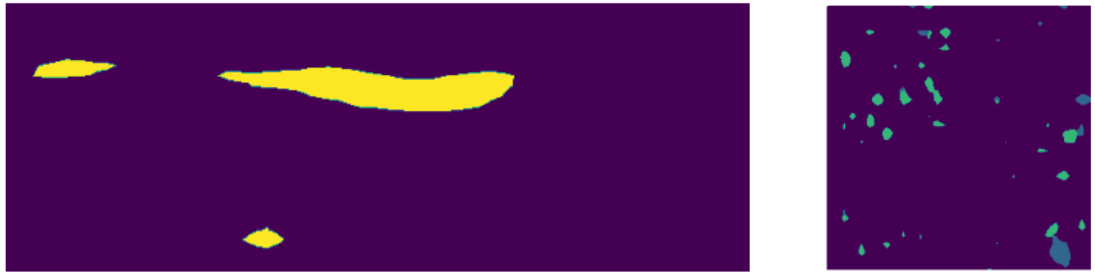}  
        \caption{SSL segmentation of small structures}
        \label{fig:mot_fault}
    \end{subfigure}
\caption{Differences in SSL segmentation depend on the nature of the target. Left: Seismic segmentation task with geological facies (large) and seismic faults (small). Right: Neuroimaging segmentation task from a mouse brain cutout, with axons (large) and cell bodies and vessels (small).}

\label{fig:motivation}
\end{figure}

Segmentation tasks provide a natural testbed for this question. In many domains, targets vary drastically in their spatial scale: large and continuous regions such as organs \cite{RAYED2024101504} or geological facies \cite{alaudah2019machine} contrast sharply with small and rare structures such as neurons, blood vessels \cite{quesada2022mtneuro}, or seismic faults \cite{prabhushankar2024cracks}. Standard SSL approaches often underperform on the latter, as pretraining objectives dominated by large contextual patterns fail to emphasize the fine-scale cues required for accurate segmentation. Figure \ref{fig:motivation} illustrates this mismatch using segmentation tasks from geophysics \cite{prabhushankar2024cracks} and neuroscience \cite{quesada2022mtneuro}: the same SSL pipeline that yields useful representations for large structures struggles to capture narrow or localized ones, either by merging them into a single large object losing granularity (seismic case), or by capturing only the most salient ones (neuroscience case).

In this work, we argue that the effectiveness of SSL in segmentation depends critically on the scale and sparsity of the target objects. To address this, we propose a simple yet effective scale-aware strategy that embeds small-window cropping directly into the augmentation process, forcing encoders to “zoom in” on localized features during pretraining. This modification biases representations toward fine-grained structures without requiring architectural changes or additional supervision. We validate this approach across two distinct scientific imaging domains: (i) seismic imaging, where faults appear as narrow, discontinuous anomalies within large volumes, and (ii) neuroimaging, where cells and other microscopic structures are small and spatially sparse relative to the surrounding tissue. Our experiments demonstrate that scale-aware SSL provides consistent improvements in both settings under limited labels, yielding up to 13\% accuracy gains for fault segmentation and 5\% for blood cell and vessel delineation. By contrast, tasks involving larger-scale features (such as seismic facies or axon regions) show little benefit, underscoring the importance of tailoring SSL strategies to the intrinsic properties of the segmentation target.
\section{Related Work}
\label{sec:related_work}

Segmentation tasks vary widely in terms of the spatial scale and density of the target structures. Deep learning has achieved strong results for segmenting large to mid-size homogeneous regions such as organs or geological facies \cite{alaudah2019machine, quesada2025large}. In contrast, small or sparse structures continue to be more challenging because of their limited spatial footprint, sparsity,  and irregular geometry \cite{karimijafarbigloo_self-supervised_2023, liu_rethinking_2024}. This gap suggests that scale strongly influences representation quality for downstream segmentation tasks.

SSL has made substantial progress with global contrastive methods (SimCLR \cite{chen2020simple}, MoCo \cite{he2020momentum}), non-contrastive ones (BYOL \cite{grill2020bootstrap}, SimSiam\cite{chen2021exploring}), among others for a wide variety of downstream tasks. However, these methods typically assume (or are optimized for) features that capture broad semantic structure. Thus many SSL pipelines achieve good transfer performance on downstream dense prediction tasks when the objects of interest have well-defined and reasonably large spatial extent, as we showcase in Figure \ref{fig:motivation}.

Recent works, however, begin to challenge this assumption. \cite{akilan2025self} observes that methods that include multiscale views, patch-based tasks, or dense pixel/patch clustering often behave better when the downstream task involves fine-grained or spatially localized structures. \cite{bardes2022vicregl} learns both global and local features to target segmentation tasks, \cite{karimijafarbigloo_self-supervised_2023} enforces inter-scale consistency, and \cite{liu_rethinking_2024} addresses inconsistencies in non-salient regions. These methods highlight the importance of balancing local and global cues during SSL pretraining.

While prior work recognizes the value of multi-scale or patch-level cues, few studies systematically analyze the impact of object scale and sparsity on SSL performance across domains. Our work addresses this gap by explicitly embedding small-window cropping into SSL pretraining and validating its effect on small, sparse targets in both seismic and neuroimaging data.
\section{Methodology}

\subsection{Self-Supervised Pretraining}
We adopt a general SSL framework where an encoder $f_{\theta}$ learns representations from unlabeled images by enforcing consistency between multiple augmented views of the same sample. Given an input image $x \in \mathbb{R}^{H \times W}$, two stochastic augmentations $t_1, t_2 \sim \mathcal{T}$ produce correlated views $x_1 = t_1(x), \; x_2 = t_2(x)$. Each view is passed through the encoder and optionally a projection head $g_{\phi}$, yielding embeddings $z_i = g_{\phi}(f_{\theta}(x_i))$. Different SSL algorithms are instantiated differently: 

\begin{itemize}
    \item \textbf{Contrastive approaches} (e.g., SimCLR, MoCo) maximize similarity between embeddings of positive pairs while contrasting them against negatives.
    \item \textbf{Non-contrastive approaches} (e.g., BYOL, SimSiam) align embeddings without explicit negatives, using asymmetry or stop-gradient mechanisms to avoid collapse.
    \item \textbf{Regularization-based approaches} (e.g., VICReg\cite{bardes2021vicreg}) enforce view alignment through an $\ell_2$ loss while explicitly preventing collapse by adding variance and covariance regularizers on the embedding distribution.
\end{itemize}

 Formally, the SSL objective can be expressed as:
\begin{equation}
    \mathcal{L}_{SSL} = \sum_{i=1}^N \ell \big( z_{i1}, z_{i2}; \{z_j\}_{j \neq i} \big),
\end{equation}
where $\ell$ denotes a contrastive, predictive, or clustering-based loss depending on the chosen SSL method, $z_{i1}, z_{i2}$ are the positive view pairs extracted from sample $x_i$, and $\{z_j\}_{j \neq i}$ is the set of negative contrastive views.

\subsubsection{Scale-Aware View Sampling}
Our key contribution is to modify the augmentation process $\mathcal{T}$ to include \textit{scale-aware cropping}, ensuring that pretraining emphasizes fine-grained patterns. Specifically, instead of restricting views to global or large crops, we explicitly sample \textit{small spatial windows} of fixed size $h \times w$ from the input $x$:
\begin{equation}
    t(x) = a(c(x)),
\end{equation}
where $c: \mathbb{R}^{H \times W} \rightarrow \mathbb{R}^{h \times w}$ is a cropping function and $a \in \mathcal{A}$ is a set of standard augmentations (flips, intensity jitter, affine transforms).

Two strategies for sampling crop centers are considered:
\begin{enumerate}
    \item \textbf{Random cropping}: patch centers are sampled uniformly from the image.
    \item \textbf{Proximity-constrained cropping}: given a first crop center $(u_1,v_1)$, a second crop is sampled within distance $\delta$, i.e.
    \begin{equation}
        \|(u_2,v_2) - (u_1,v_1)\|_2 < \delta,
    \end{equation}
    encouraging overlap and spatial coherence.
\end{enumerate}

Embedding this small-window sampling directly into SSL forces the encoder to attend to localized structures that may otherwise be underrepresented in large-scale views. This design is agnostic to the underlying SSL objective and can be plugged into contrastive, non-contrastive, or clustering-based frameworks.

\subsection{Downstream Segmentation}
After pretraining, the encoder \( f_\theta \) is integrated into a supervised segmentation network for fault delineation. We adopt a standard encoder–decoder architecture, where the pretrained encoder initializes the feature extraction layers and a randomly initialized decoder is appended to produce voxel-wise segmentation maps. During this stage, training is performed on a small labeled subset of the seismic dataset, consistent with the label-constrained regime.

Formally, given a labeled image \( x \in \mathbb{R}^{H \times W} \) and its corresponding binary fault mask \( Y \in \{0,1\}^{H \times W} \), we extract training samples as patches of the same spatial size (\( h \times w \)) used during SSL pretraining $x_{(u,v)}, y_{(u,v)}$, where \( (u,v) \) denotes the patch center coordinates. Each patch is passed through the encoder–decoder to yield a predicted mask 
\[
\hat{y}_{(u,v)} = \sigma\!\left(F_\psi(f_\theta(x_{(u,v)}))\right),
\]
with \( F_\psi \) denoting the decoder and \( \sigma \) the sigmoid activation. Training minimizes the  Dice loss between \( \hat{y}_{(u,v)} \) and \( y_{(u,v)} \).

At inference time, a full image is segmented by sliding a window of size \( h \times w \)  with stride \( s < h \). Predictions from overlapping patches are averaged to produce the stitched full-slice segmentation. This reconstruction ensures seamless predictions across patch boundaries.

Performance is evaluated by comparing stitched predictions \( \hat{Y} \) against ground-truth masks \( Y \) using standard segmentation metrics. This patch-based training and inference strategy ensures compatibility with the encoder’s pretraining resolution while simultaneously retaining the original scale of the image.

\section{Results}
\label{sec:results}

\subsection{Experimental Setup}

We evaluate our approach on two distinct domains that naturally lend themselves to segmentation of small or sparse structures:
\begin{itemize}
    \item \textbf{Seismic fault segmentation}, where targets correspond to thin and discontinuous linear faults embedded in noisy seismic volumes. We use the \texttt{CRACKS} \cite{prabhushankar2024cracks} and \texttt{Thebe} \cite{AN2021107219} datasets for this purpose.
    \item \textbf{Segmentation of cellular structures in neuroimaging data}, where targets are small, compact objects relative to surrounding tissue. We use the \texttt{MTNeuro} dataset \cite{quesada2022mtneuro} for this domain.
\end{itemize}
For each dataset, the full set of training images are available for self-supervised pretraining, while only a small fraction of labeled samples (10\% unless otherwise noted) is used for downstream segmentation. This reflects the label-constrained regime common in different domains of scientific imaging, where acquiring labels is a costly and time-consuming endeavor.

We evaluate three representative SSL methods: SimCLR, BYOL, and VICReg, covering multiple SSL paradigms. All models share a ResNet-18 backbone, trained for 100 epochs with batch size 128 and the LARS optimizer. Scale-aware cropping is embedded into the augmentation pipeline, with patches of size $L/2$, $L/4$, and $L/8$ sampled per image ($L$ is the smallest image dimension for each dataset). Both random and proximity-constrained cropping strategies are tested, combined with standard augmentations (flips, intensity jitter, affine transforms). Results are reported as averages across the three SSL methods to reduce method-specific variance.

We compare against fully supervised models trained directly on the limited labeled data, and SSL baselines including vanilla SimCLR trained with global scope only (no scale-aware sampling) and VICRegL \cite{bardes2022vicregl}, a VICReg adaptation targeted towards segmentation by focusing on global and local features. For downstream segmentation, pretrained encoders are coupled with a DeepLabV3 decoder and fine-tuned using Dice loss. Inference is performed with overlapping sliding windows. Performance is evaluated with Dice coefficient (region overlap) and Hausdorff distance (structural and boundary alignment).

To complement our experiments, we also leverage the fact that both the \texttt{CRACKS} and \texttt{MTNeuro} datasets provide large-structure segmentation targets. On \texttt{CRACKS}, we evaluate facies segmentation, where lithostratigraphic units occupy a much larger portion of the seismic volume than faults. On \texttt{MTNeuro}, we evaluate axon segmentation, targeting neural structures substantially larger than the cells and vessels used in our main setup. These additional tasks allow us to assess how the \textit{same SSL representations} transfer to segmentation targets of different scales within the same data.  

All SSL pretraining experiments were conducted on a single NVIDIA GTX 3060 GPU with an Intel i7-6700K CPU. Scale-aware SSL replaces full-resolution views with small-window crops, reducing effective input resolution during pretraining. As a result, SSL pretraining with scale-aware cropping is up to 3× faster than full-image SSL using the same backbone, optimizer, and batch size, while also reducing memory usage and requiring no architectural changes.

\begin{table*}[t]
\centering
\caption{Experiments across seismic datasets}
\label{tab:seismic_results}
\resizebox{\textwidth}{!}{
\begin{tabular}{c||cccccc||c||cc|cc|cc}
\multirow{3}{*}{Dataset} & \multicolumn{6}{c||}{Full-slice Baselines} & \multirow{3}{*}{SSL Patch Methods} & \multicolumn{2}{c|}{\multirow{2}{*}{Patch size L/2}} & \multicolumn{2}{c|}{\multirow{2}{*}{Patch size L/4}} & \multicolumn{2}{c}{\multirow{2}{*}{Patch size L/8}} \\ \cline{2-7}
 & \multicolumn{2}{c|}{Supervised} & \multicolumn{2}{c|}{SSL} & \multicolumn{2}{c||}{VICRegL} &  & \multicolumn{2}{c|}{} & \multicolumn{2}{c|}{} & \multicolumn{2}{c}{} \\ \cline{2-7} \cline{9-14}
 & HD & \multicolumn{1}{c|}{Dice} & HD & \multicolumn{1}{c|}{Dice} & HD & Dice &  & HD & Dice & HD & Dice & HD & Dice \\ \hline
\multirow{2}{*}{\texttt{CRACKS}} 
 & \multirow{2}{*}{95.51} & \multicolumn{1}{c|}{\multirow{2}{*}{0.600}} 
 & \multirow{2}{*}{88.42} & \multicolumn{1}{c|}{\multirow{2}{*}{0.591}} 
 & \multirow{2}{*}{61.69} & \multirow{2}{*}{0.615} 
 & Random   & 14.27 & 0.619 & 11.92 & 0.615 & \cellcolor[HTML]{67FD9A}10.35 & \cellcolor[HTML]{67FD9A}0.625 \\
 &  & \multicolumn{1}{c|}{} &  & \multicolumn{1}{c|}{} &  &  
 & Distance & 14.51 & 0.615 & 14.09 & 0.601 & \cellcolor[HTML]{67FD9A}10.15 & \cellcolor[HTML]{67FD9A}0.631 \\ \hline
\multirow{2}{*}{\texttt{Thebe}} 
 & \multirow{2}{*}{110.39} & \multicolumn{1}{c|}{\multirow{2}{*}{0.513}} 
 & \multirow{2}{*}{200} & \multicolumn{1}{c|}{\multirow{2}{*}{0.491}} 
 & \multirow{2}{*}{36.08} & \multirow{2}{*}{0.507} 
 & Random   & 62.95 & 0.517 & 14.32 & 0.582 & \cellcolor[HTML]{67FD9A}13.46 & \cellcolor[HTML]{67FD9A}0.612 \\
 &  & \multicolumn{1}{c|}{} &  & \multicolumn{1}{c|}{} &  &  
 & Distance & 33.71 & 0.548 & 20.15 & 0.563 & \cellcolor[HTML]{67FD9A}13.75 & \cellcolor[HTML]{67FD9A}0.591 \\ \hline\hline
\multirow{2}{*}{\begin{tabular}[c]{@{}c@{}}\texttt{CRACKS}\\(facies)\end{tabular}}
 & \multirow{2}{*}{0.027} & \multicolumn{1}{c|}{\multirow{2}{*}{0.793}} 
 & \multirow{2}{*}{0.029} & \multicolumn{1}{c|}{\multirow{2}{*}{0.792}} 
 & \multirow{2}{*}{0.18} & \multirow{2}{*}{0.662} 
 & Random   & \cellcolor[HTML]{67FD9A}0.019 & \cellcolor[HTML]{67FD9A}0.871 & 0.178 & 0.763 & 0.618 & 0.702 \\
 &  & \multicolumn{1}{c|}{} &  & \multicolumn{1}{c|}{} &  &  
 & Distance & \cellcolor[HTML]{67FD9A}0.029 & \cellcolor[HTML]{67FD9A}0.863 & 0.086 & 0.802 & 0.821 & 0.694 \\ \hline
\end{tabular}
}
\end{table*}

\begin{table*}[t]
\centering
\caption{Experiments across neuroimaging data}
\label{tab:neuro_results}
\resizebox{\textwidth}{!}{
\begin{tabular}{c||cccccc||c||cc|cc|cc}
 \multirow{3}{*}{\shortstack{Neural\\Segmentation\\Target}} & \multicolumn{6}{c||}{Full-slice Baselines} & \multirow{3}{*}{SSL Patch Methods} & \multicolumn{2}{c|}{\multirow{2}{*}{Patch size L/2}} & \multicolumn{2}{c|}{\multirow{2}{*}{Patch size L/4}} & \multicolumn{2}{c}{\multirow{2}{*}{Patch size L/8}} \\ \cline{2-7}
 & \multicolumn{2}{c|}{Supervised} & \multicolumn{2}{c|}{SSL} & \multicolumn{2}{c||}{VICRegL} &  & \multicolumn{2}{c|}{} & \multicolumn{2}{c|}{} & \multicolumn{2}{c}{} \\ \cline{2-7} \cline{9-14}
 & HD & \multicolumn{1}{c|}{Dice} & HD & \multicolumn{1}{c|}{Dice} & HD & Dice &  & HD & Dice & HD & Dice & HD & Dice \\ \hline
\multirow{2}{*}{\begin{tabular}[c]{@{}c@{}}Cells\&Vessels\\ (small)\end{tabular}}
 & \multirow{2}{*}{1.85} & \multicolumn{1}{c|}{\multirow{2}{*}{0.686}}
 & \multirow{2}{*}{1.78} & \multicolumn{1}{c|}{\multirow{2}{*}{0.685}}
 & \multirow{2}{*}{48.99} & \multirow{2}{*}{0.340}
 & Random   & 1.16 & 0.736 & 1.13 & 0.731 & \cellcolor[HTML]{67FD9A}1.16 & \cellcolor[HTML]{67FD9A}0.733 \\
 &  & \multicolumn{1}{c|}{} &  & \multicolumn{1}{c|}{} &  & 
 & Distance & 1.18 & 0.736 & 1.18 & 0.734 & \cellcolor[HTML]{67FD9A}1.30 & \cellcolor[HTML]{67FD9A}0.725 \\ \hline\hline
\multirow{2}{*}{\begin{tabular}[c]{@{}c@{}}Axons\\ (large)\end{tabular}}
 & \multirow{2}{*}{41.25} & \multicolumn{1}{c|}{\multirow{2}{*}{0.805}}
 & \multirow{2}{*}{43.02} & \multicolumn{1}{c|}{\multirow{2}{*}{0.803}}
 & \multirow{2}{*}{29.84} & \multirow{2}{*}{0.589}
 & Random   & \cellcolor[HTML]{67FD9A}25.06 & \cellcolor[HTML]{67FD9A}0.809 & 23.30 & 0.800 & 24.39 & 0.790 \\
 &  & \multicolumn{1}{c|}{} &  & \multicolumn{1}{c|}{} &  & 
 & Distance & \cellcolor[HTML]{67FD9A}20.83 & \cellcolor[HTML]{67FD9A}0.817 & 24.52 & 0.814 & 23.27 & 0.796 \\ \hline
\end{tabular}
}
\end{table*}

\subsection{Results and analysis}

\begin{figure}[t]
    \begin{subfigure}{\linewidth}
    \centering
        \includegraphics[width=\textwidth]{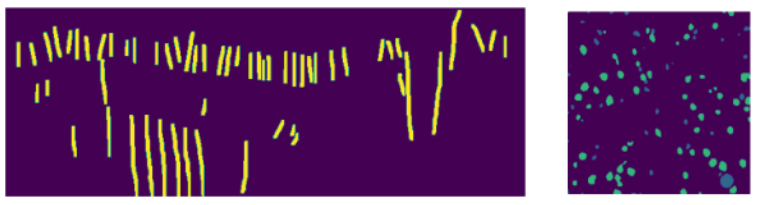} 
        \caption{Ground truth}
        \label{fig:res_gt}
    \end{subfigure}
    \begin{subfigure}{\linewidth}
        \includegraphics[width=\textwidth]{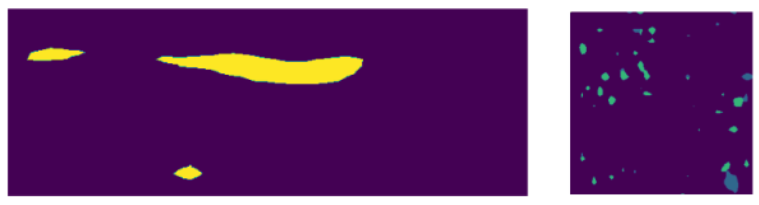}  
        \caption{Standard SSL Segmentation (SimCLR)}
        \label{fig:res_ssl}
    \end{subfigure}
    \begin{subfigure}{\linewidth}
        \includegraphics[width=\textwidth]{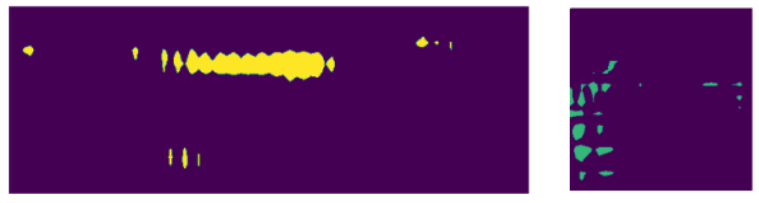}  
        \caption{Multi-crop SSL segmentation (VICRegL)}
        \label{fig:res_vic}
    \end{subfigure}
    \begin{subfigure}{\linewidth}
        \includegraphics[width=\textwidth]{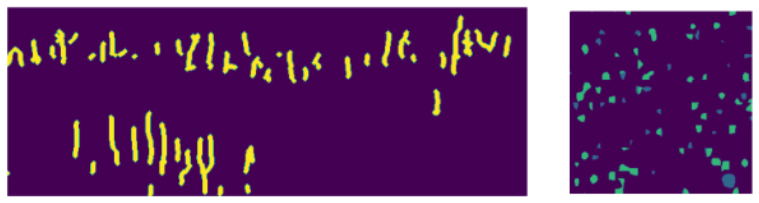}  
        \caption{Scale-aware SSL segmentation (proposed)}
        \label{fig:res_patch}
    \end{subfigure}
\caption{Qualitative segmentation results for \texttt{CRACKS} (left) and \texttt{MTNeuro} (right).}
\label{fig:results}
\end{figure}

We showcase our results across both domains in Tables~\ref{tab:seismic_results} and~\ref{tab:neuro_results}, where the best performing setup for each dataset is highlighted in green. In both cases, the left-hand side of the tables contains results for the full-slice baselines (including the multi-crop VICRegL \cite{bardes2022vicregl}) and the right-hand side contains our patch-based experiments. 

\textbf{Metric interpretation.} Dice score measures pixel-level overlap between prediction and ground truth, emphasizing volumetric accuracy, while Hausdorff distance highlights boundary alignment and structural fidelity. Reporting both allows us to capture not only whether objects are segmented at all, but also whether their geometry is reconstructed correctly.

\textbf{Small-structure segmentation.} We observe that scale-aware SSL with smaller patch sizes consistently improves performance when the targets are small or sparse. For seismic fault segmentation (Table \ref{tab:seismic_results}, first and second row), Dice scores increase by up to 10\% compared to global-view SSL, with corresponding reductions in Hausdorff distance. For cell and vessel segmentation in neuroimaging, Dice scores improve by up to 5\% over the supervised baseline. It is also worth noting that the Hausdorff distance decreases most drastically in the seismic experiments, indicating a much better structural alignment of the detected faults (as showcased visually in Figure \ref{fig:results}).  
Furthermore, in these tasks the smallest patches ($L/8$) yield the strongest gains, confirming that aggressively zooming in on local regions provides the best inductive bias for thin or compact structures. Importantly, our approach outperforms multi-crop VICRegL (a state-of-the-art method designed explicitly for segmentation) particularly in the small-structure experiments, demonstrating that direct scale-aware cropping is a stronger bias than generic multi-crop augmentations.

\textbf{Large-structure segmentation.} In contrast, the same representations do not yield significant improvements for larger structures, as seen in the bottom rows of Tables~\ref{tab:seismic_results} and~\ref{tab:neuro_results}. For facies segmentation in seismic volumes and axon segmentation in neuroimaging, smaller patches provide little to no benefit, and performance consistently degrades as patch size decreases. In these cases, the largest patches ($L/2$) still provide modest improvements over the full-slice baselines, since segmenting large continuous regions requires retaining broader contextual cues that are discarded when sampling very small windows. This discrepancy underscores that the effectiveness of scale-aware SSL depends critically on the spatial footprint of the target.

\textbf{Qualitative analysis.} Figure~\ref{fig:results} further illustrates these trends. While multi-crop VICRegL (Figure \ref{fig:res_vic}) captures more granular structures than vanilla SSL methods, it still fails to reproduce many of the fine-scale discontinuities in faults or the detailed boundaries of cells. By contrast, our patch-based strategy (Figure \ref{fig:res_patch}) recovers significantly more local structure, yielding boundaries that align closely with the ground truth. This qualitative gap reinforces the quantitative improvements observed in Dice and Hausdorff scores.

Taken together, results consistently reveal that SSL representations are not universally effective: their utility depends critically on the spatial footprint of the target. This scale-dependence emerges as a governing factor that determines whether SSL pretraining provides benefits or fails to transfer. Our proposed scale-aware cropping strategy acts as a strong inductive bias for small, sparse structures but can hinder performance on larger, homogeneous ones. This is because smaller crops emphasize localized discontinuities and fine-scale patterns essential for thin or compact objects, but overlook global context that is critical for segmenting extended regions.

\section{Conclusion}

To overcome the inductive biases traditional SSL methods have toward large or dominant structures, we introduce a scale-aware SSL strategy that embeds small-window cropping into the augmentation process, enabling encoders to better capture fine-grained structures during pretraining. Across seismic and neuroimaging domains, our experiments consistently show that this approach yields substantial gains for segmenting small or sparse targets such as faults and cells, while offering limited or even negative benefit for larger structures such as facies and axons. These findings reveal that the effectiveness of SSL is fundamentally scale-dependent: the utility of learned representations is governed by the spatial footprint of the target objects. Rather than being a universal recipe, SSL must be aligned with the geometry of the downstream task. This principle motivates future research into adaptive multi-scale strategies that can reconcile both small and large regimes, ensuring that SSL pipelines remain effective across the full spectrum of scientific imaging tasks.




\bibliographystyle{IEEEbib}
\bibliography{ref}

@inproceedings{chen2020simple,
  title={A simple framework for contrastive learning of visual representations},
  author={Chen, Ting and Kornblith, Simon and Norouzi, Mohammad and Hinton, Geoffrey},
  booktitle={International conference on machine learning},
  pages={1597--1607},
  year={2020},
  organization={PmLR}
}

@inproceedings{he2020momentum,
  title={Momentum contrast for unsupervised visual representation learning},
  author={He, Kaiming and Fan, Haoqi and Wu, Yuxin and Xie, Saining and Girshick, Ross},
  booktitle={Proceedings of the IEEE/CVF conference on computer vision and pattern recognition},
  pages={9729--9738},
  year={2020}
}

@inproceedings{yeh2022decoupled,
  title={Decoupled contrastive learning},
  author={Yeh, Chun-Hsiao and Hong, Cheng-Yao and Hsu, Yen-Chi and Liu, Tyng-Luh and Chen, Yubei and LeCun, Yann},
  booktitle={European conference on computer vision},
  pages={668--684},
  year={2022},
  organization={Springer}
}

@article{grill2020bootstrap,
  title={Bootstrap your own latent-a new approach to self-supervised learning},
  author={Grill, Jean-Bastien and Strub, Florian and Altch{\'e}, Florent and Tallec, Corentin and Richemond, Pierre and Buchatskaya, Elena and Doersch, Carl and Avila Pires, Bernardo and Guo, Zhaohan and Gheshlaghi Azar, Mohammad and others},
  journal={Advances in neural information processing systems},
  volume={33},
  pages={21271--21284},
  year={2020}
}

@inproceedings{chen2021exploring,
  title={Exploring simple siamese representation learning},
  author={Chen, Xinlei and He, Kaiming},
  booktitle={Proceedings of the IEEE/CVF conference on computer vision and pattern recognition},
  pages={15750--15758},
  year={2021}
}

@article{quesada2022mtneuro,
  title={Mtneuro: A benchmark for evaluating representations of brain structure across multiple levels of abstraction},
  author={Quesada, Jorge and Sathidevi, Lakshmi and Liu, Ran and Ahad, Nauman and Jackson, Joy and Azabou, Mehdi and Xiao, Jingyun and Liding, Christopher and Jin, Matthew and Urzay, Carolina and others},
  journal={Advances in neural information processing systems},
  volume={35},
  pages={5299--5314},
  year={2022}
}

@inproceedings{li2024self,
  title={Self-supervised alignment learning for medical image segmentation},
  author={Li, Haofeng and Ouyang, Yiming and Wan, Xiang},
  booktitle={2024 IEEE International Symposium on Biomedical Imaging (ISBI)},
  pages={1--5},
  year={2024},
  organization={IEEE}
}

@article{wang2022self,
  title={Self-supervised learning in remote sensing: A review},
  author={Wang, Yi and Albrecht, Conrad M and Braham, Nassim Ait Ali and Mou, Lichao and Zhu, Xiao Xiang},
  journal={IEEE Geoscience and Remote Sensing Magazine},
  volume={10},
  number={4},
  pages={213--247},
  year={2022},
  publisher={IEEE}
}

@article{muhtar2023cmid,
  title={Cmid: A unified self-supervised learning framework for remote sensing image understanding},
  author={Muhtar, Dilxat and Zhang, Xueliang and Xiao, Pengfeng and Li, Zhenshi and Gu, Feng},
  journal={IEEE Transactions on Geoscience and Remote Sensing},
  volume={61},
  pages={1--17},
  year={2023},
  publisher={IEEE}
}

@article{RAYED2024101504,
title = {Deep learning for medical image segmentation: State-of-the-art advancements and challenges},
journal = {Informatics in Medicine Unlocked},
volume = {47},
pages = {101504},
year = {2024},
issn = {2352-9148},
doi = {https://doi.org/10.1016/j.imu.2024.101504},
url = {https://www.sciencedirect.com/science/article/pii/S2352914824000601},
author = {Md. Eshmam Rayed and S.M. Sajibul Islam and Sadia Islam Niha and Jamin Rahman Jim and Md Mohsin Kabir and M.F. Mridha}
}

@article{alaudah2019machine,
  title={A machine-learning benchmark for facies classification},
  author={Alaudah, Yazeed and Micha{\l}owicz, Patrycja and Alfarraj, Motaz and AlRegib, Ghassan},
  journal={Interpretation},
  volume={7},
  number={3},
  pages={SE175--SE187},
  year={2019},
  publisher={Society of Exploration Geophysicists and American Association of Petroleum~…}
}

@article{prabhushankar2024cracks,
  title={CRACKS: Crowdsourcing Resources for Analysis and Categorization of Key Subsurface faults},
  author={Prabhushankar, Mohit and Kokilepersaud, Kiran and Quesada, Jorge and Yarici, Yavuz and Zhou, Chen and Alotaibi, Mohammad and AlRegib, Ghassan and Mustafa, Ahmad and Kumakov, Yusufjon},
  journal={arXiv preprint arXiv:2408.11185},
  year={2024}
}

@misc{karimijafarbigloo_self-supervised_2023,
	title = {Self-supervised {Semantic} {Segmentation}: {Consistency} over {Transformation}},
	shorttitle = {Self-supervised {Semantic} {Segmentation}},
	url = {http://arxiv.org/abs/2309.00143},
	doi = {10.48550/arXiv.2309.00143},
	urldate = {2024-11-26},
	publisher = {arXiv},
	author = {Karimijafarbigloo, Sanaz and Azad, Reza and Kazerouni, Amirhossein and Velichko, Yury and Bagci, Ulas and Merhof, Dorit},
	month = aug,
	year = {2023},
	note = {arXiv:2309.00143 [cs]},
}

@article{liu_rethinking_2024,
	title = {Rethinking {Self}-{Supervised} {Semantic} {Segmentation}: {Achieving} {End}-to-{End} {Segmentation}},
	volume = {46},
	copyright = {https://creativecommons.org/licenses/by/4.0/legalcode},
	issn = {0162-8828, 2160-9292, 1939-3539},
	shorttitle = {Rethinking {Self}-{Supervised} {Semantic} {Segmentation}},
	url = {https://ieeexplore.ieee.org/document/10607955/},
	doi = {10.1109/TPAMI.2024.3432326},
	number = {12},
	urldate = {2024-11-26},
	journal = {IEEE Transactions on Pattern Analysis and Machine Intelligence},
	author = {Liu, Yue and Zeng, Jun and Tao, Xingzhen and Fang, Gang},
	month = dec,
	year = {2024},
	pages = {10036--10046},
}

@article{quesada2025large,
  title={A Large-scale Benchmark on Geological Fault Delineation Models: Domain Shift, Training Dynamics, Generalizability, Evaluation and Inferential Behavior},
  author={Quesada, Jorge and Zhou, Chen and Chowdhury, Prithwijit and Alotaibi, Mohammad and Mustafa, Ahmad and Kumamnov, Yusufjon and Prabhushankar, Mohit and AlRegib, Ghassan},
  journal={arXiv preprint arXiv:2505.08585},
  year={2025}
}

@article{akilan2025self,
  title={Self-Supervised Learning for Image Segmentation: A Comprehensive Survey},
  author={Akilan, Thangarajah and Jahan, Nusrat and Zhang, Wandong},
  journal={arXiv preprint arXiv:2505.13584},
  year={2025}
}

@article{AN2021107219,
title = {A gigabyte interpreted seismic dataset for automatic fault recognition},
journal = {Data in Brief},
volume = {37},
pages = {107219},
year = {2021},
issn = {2352-3409},
doi = {https://doi.org/10.1016/j.dib.2021.107219},
url = {https://www.sciencedirect.com/science/article/pii/S2352340921005035},
author = {Yu An and Jiulin Guo and Qing Ye and Conrad Childs and John Walsh and Ruihai Dong},
}

@article{bardes2021vicreg,
  title={Vicreg: Variance-invariance-covariance regularization for self-supervised learning},
  author={Bardes, Adrien and Ponce, Jean and LeCun, Yann},
  journal={arXiv preprint arXiv:2105.04906},
  year={2021}
}

@article{bardes2022vicregl,
  title={Vicregl: Self-supervised learning of local visual features},
  author={Bardes, Adrien and Ponce, Jean and LeCun, Yann},
  journal={Advances in Neural Information Processing Systems},
  volume={35},
  pages={8799--8810},
  year={2022}
}

@inproceedings{devlin2019bert,
  title={Bert: Pre-training of deep bidirectional transformers for language understanding},
  author={Devlin, Jacob and Chang, Ming-Wei and Lee, Kenton and Toutanova, Kristina},
  booktitle={Proceedings of the 2019 conference of the North American chapter of the association for computational linguistics: human language technologies, volume 1 (long and short papers)},
  pages={4171--4186},
  year={2019}
}

@inproceedings{caron2021emerging,
  title={Emerging properties in self-supervised vision transformers},
  author={Caron, Mathilde and Touvron, Hugo and Misra, Ishan and J{\'e}gou, Herv{\'e} and Mairal, Julien and Bojanowski, Piotr and Joulin, Armand},
  booktitle={Proceedings of the IEEE/CVF international conference on computer vision},
  pages={9650--9660},
  year={2021}
}

@InProceedings{Quesada_2024_CVPR,
    author    = {Quesada, Jorge and Alotaibi, Mohammad and Prabhushankar, Mohit and Alregib, Ghassan},
    title     = {PointPrompt: A Multi-modal Prompting Dataset for Segment Anything Model},
    booktitle = {Proceedings of the IEEE/CVF Conference on Computer Vision and Pattern Recognition (CVPR) Workshops},
    month     = {June},
    year      = {2024},
    pages     = {1604-1610}
}

@inproceedings{quesada2024benchmarking,
  title={Benchmarking Human and Automated Prompting in the Segment Anything Model},
  author={Quesada, Jorge and Fowler, Zoe and Alotaibi, Mohammad and Prabhushankar, Mohit and AlRegib, Ghassan},
  booktitle={2024 IEEE International Conference on Big Data (BigData)},
  pages={1625--1634},
  year={2024},
  organization={IEEE}
}

@inproceedings{kirillov2023segment,
  title={Segment anything},
  author={Kirillov, Alexander and Mintun, Eric and Ravi, Nikhila and Mao, Hanzi and Rolland, Chloe and Gustafson, Laura and Xiao, Tete and Whitehead, Spencer and Berg, Alexander C and Lo, Wan-Yen and others},
  booktitle={Proceedings of the IEEE/CVF International Conference on Computer Vision},
  pages={4015--4026},
  year={2023}
}



\end{document}